\documentclass[sigconf,natbib=true,anonymous=false]{acmart}
\AtBeginDocument{%
  \providecommand\BibTeX{{%
    \normalfont B\kern-0.5em{\scshape i\kern-0.25em b}\kern-0.8em\TeX}}}

\setcopyright{acmcopyright}
\copyrightyear{2024}
\acmYear{2024}
\acmDOI{XXXXXXX.XXXXXXX}

\acmConference[Conference acronym 'XX]{}{June 03--05,
  2018}{Woodstock, NY}

\usepackage{pdflscape}
\usepackage{afterpage}
\usepackage{soul}
\usepackage{caption}
\usepackage{amsmath}
\usepackage{amsthm}
\usepackage{booktabs}
\usepackage{algorithm}
\usepackage{algorithmic}
\usepackage{colortbl}
\usepackage{textcomp}
\usepackage{arydshln}
\usepackage{array,multirow,graphicx}
\usepackage{xcolor} 
\usepackage{makecell}
\usepackage{soul}
\usepackage{tikz}
\usetikzlibrary{positioning,fit,arrows.meta,backgrounds}
\usetikzlibrary{arrows.meta,
                chains,
                positioning,
                shapes.geometric
                }

\usepackage{graphicx}
\usepackage{booktabs}
\usepackage{tikz}
\usetikzlibrary{trees}

\tikzset{
    module/.style={%
        draw, rounded corners,
        minimum width=#1,
        minimum height=7mm,
        font=\sffamily
        },
    module/.default=2cm,
    >=LaTeX
}
\tikzset{
    module-1/.style={%
        minimum width=10mm,
        minimum height=7mm,
        font=\sffamily
        },
    module-1/.default=2cm,
    >=LaTeX
}

\tikzset{
    module2/.style={%
        draw, rounded corners,
        minimum width=#1,
        minimum height=7mm,
        align=center,
        font=\sffamily
        },
    module2/.default=2cm,
    >=LaTeX
}

\tikzset{
    module3/.style={%
        minimum width=10mm,
        minimum height=7mm,
        align=center,
        font=\sffamily
        },
    module-1/.default=2cm,
    >=LaTeX
}

\tikzset{
    module4/.style={%
        draw, rounded corners,
        minimum width=46mm,
        align=center,
        minimum height=15mm,
        font=\sffamily
        },
    module/.default=2cm,
    >=LaTeX
}

\tikzstyle{pinstyle} = [pin edge={<-,ultra thick,black}]
\tikzstyle{pinstyle-1} = [pin edge={->,ultra thick,black}]
\tikzstyle{pinstyle-2} = [pin distance=9cm, edge={->,ultra thick,black}]

\begin{document}


\title{MultiLS: An End-to-End Lexical Simplification Framework}

\title{MultiLS: A Multi-task Lexical Simplification Framework}




\author{Kai North}
\affiliation{George Mason University
\country{USA}}
\email{knorth8@gmu.edu}

\author{Tharindu Ranasinghe} 
\affiliation{Aston University
\country{UK}}
\email{t.ranasinghe@aston.ac.uk}

\author{Matthew Shardlow}
\affiliation{Manchester Metropolitan University
\country{UK}}
\email{m.shardlow@mmu.ac.uk}

\author{Marcos Zampieri} 
\affiliation{George Mason University
\country{USA}}
\email{mzampier@gmu.edu}

\renewcommand{\shortauthors}{North, et al.}



\begin{abstract}
Lexical Simplification (LS) automatically replaces difficult to read words for easier alternatives while preserving a sentence’s original meaning. LS is a precursor to Text Simplification with the aim of improving text accessibility to various target demographics, including children, second language learners, individuals with reading disabilities or low literacy. Several datasets exist for LS. These LS datasets specialize on one or two sub-tasks within the LS pipeline.  However, as of this moment, no single LS dataset has been developed that covers all LS sub-tasks. We present MultiLS, the first LS framework that allows for the creation of a multi-task LS dataset. We also present MultiLS-PT, the first dataset to be created using the MultiLS framework. We demonstrate the potential of MultiLS-PT by carrying out all LS sub-tasks of (1). lexical complexity prediction (LCP), (2). substitute generation, and (3). substitute ranking for Portuguese. Model performances are reported, ranging from transformer-based models to more recent large language models (LLMs).
\end{abstract}





\maketitle

\section{Introduction}
Lexical Simplification (LS) makes jargon-rich texts instantly accessible to a low-literacy population \cite{LCPsurvey}. LS replaces complex words with simpler alternatives while maintaining the sentence's original meaning: 

\begin{center}
``Seek \textbf{\textit{consultation}} for \textbf{\textit{diagnosis}}'' $\rightarrow$ ``Seek \textbf{\textit{advice}} for \textbf{\textit{illness}}''

\end{center}

\noindent Despite the growing popularity of LS \cite{paetzold-specia:2016:SemEval1, yiman-EtAl:2018:BEA, semeval-2021, tsar2022}, there exist few datasets for the full training and evaluation of LS systems. In fact, all publicly available datasets, regardless of language, fail to cover all sub-tasks within the LS pipeline: lexical complexity prediction (LCP), substitute generation (SG), selection (SS), and ranking (SR) \cite{maddela2018word, shardlow-etal-2020-complex}. 

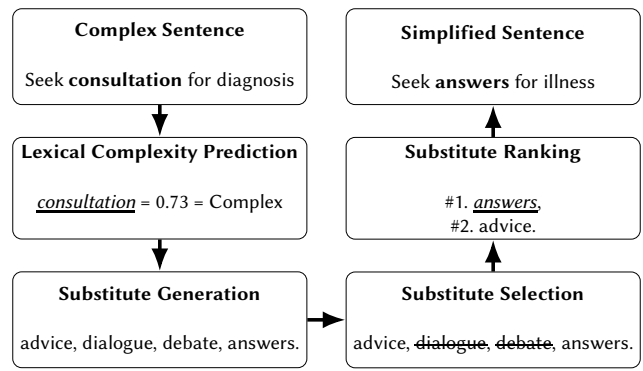
\begin{figure}[!ht]
  \centering
\scalebox{0.85}{\begin{tikzpicture}
\node[module4, fill=white!5] (I1) {\textbf{Complex Sentence}\\\\ 
Seek \textbf{consultation} for diagnosis};
\node[module4, below=0.5cm of I1] (I2) {\textbf{Lexical Complexity Prediction}\\\\\underline{\textit{consultation}} = 0.73 = Complex \\} ;
\node[module4, fill=white!5, below=0.5cm of I2] (I3) {\textbf{Substitute Generation}\\\\advice, dialogue, debate, answers.};
\draw[->, ultra thick] (I1)--(I2);
\draw[->, ultra thick] (I2)--(I3);

\node[module4, fill=white!5, right=0.6cm of I1] (I6) {\textbf{Simplified Sentence}\\\\Seek \textbf{answers} for illness};
\node[module4, fill=white!5, below=0.5cm of I6] (I5) {\textbf{Substitute Ranking}\\\\ \#1. \underline{\textit{answers}}, \\\#2. advice.};
\node[module4, fill=white!5, below=0.5cm of I5] (I4) {\textbf{Substitute Selection}\\\\ advice, \st{dialogue}, \st{debate}, answers.};
\draw[->, ultra thick] (I3)--(I4);
\draw[->, ultra thick] (I4)--(I5);
\draw[->, ultra thick] (I5)--(I6);
\end{tikzpicture}}
\caption{LS Pipeline. Example shows LS pipeline applied within the biomedical domain. Original figure adapted from \cite{Paetzold2015}}.
\label{pipeline_figure}
\end{figure}

Previous end-to-end LS frameworks, such as BenchLS \cite{Paetzold2016a}, SIMPLEX PB 3.0 \cite{Hartmann2020}, TSAR-2022 \cite{tsar2022}, and others \cite{mccarthy-navigli-2007-semeval, specia2012, Horn2014, Paetzold2016, Alarcn2021LexicalSS, HanLS}, have collected gold simplifications needed for substitute generation, selection, and ranking yet have excluded the lexical complexities required for LCP. In contrast, others have gathered lexical complexities but have refrained from collecting gold simplifications, such as CWI-2018 \cite{yiman-EtAl:2018:BEA}, WCL \cite{maddela2018word}, and LCP-2021 \cite{shardlow-etal-2020-complex}. \textcolor{black}{Each of these LS frameworks also annotated different target words meaning that their subsequent datasets cannot be combined to provide all the necessary information for LS.} In addition, only a few datasets contained texts from several genres \cite{specia2012, kajiwara-yamamoto-2015-evaluation, Paetzold2016a, Lee_Yeung2018, shardlow-etal-2020-complex}. Therefore, no previous LS dataset can be used for the full training and evaluation of an entire LS system, from LCP to substitute generation, selection, and ranking as depicted in Figure \ref{pipeline_figure}.

\noindent We introduce MultiLS, the first multi-purpose, and multi-genre framework for the creation of an all-in-one style LS dataset. MultiLS is the first LS framework to provide target words with lexical complexity values required for LCP and gold candidate simplifications needed for substitute generation, selection, and ranking. MultiLS is an extensible framework allowing for datasets in various languages containing texts from domains to be constructed using the same annotation framework. This opens exciting possibilities on the use of domain adaptation techniques, cross-lingual learning, and multi-task learning to simplification. MultiLS contributes to advancing the state-of-the-art on text simplification which is a vital technology for accessibility. Simplification is a topic of extreme relevance to the web and information retrieval communities as evidenced by multiple papers on the topic published at SIGIR in the past few years \cite{Pattisapu_etal2020, Casola_etal2023}. 


We use MultiLS to create MultiLS-PT, the first multi-task and multi-genre dataset for Portuguese LS. Portuguese is one of the ten most spoken languages in the world with over 250 million speakers \cite{eberhard2023}. Many countries where Portuguese is spoken such as Angola, Brazil, and Mozambique have low literacy rates. We chose to include texts from the Brazilian variety in MultiLS-PT as this is the most widely-spoken variety of Portuguese. Furthermore, while Brazil is one of the largest economies in the world, a large part of its population, predominantly in the north and northeast regions of the country, are either illiterate or functionally illiterate worsening existing socio-economic challenges \cite{ireland2008}. Research shows that only 1\% of Brazil's rural population are rated as fully proficient readers hindering their socio-economic mobility \cite{leal-etal-2018-nontrivial}. As such, there is ample motivation for the development of assistive reading technologies for Portuguese which inspired us to create MultiLS-PT.

MultiLS-PT is the first Portuguese LS dataset to contain instances from the Bible, news articles, and biomedical texts. MultiLS-PT is also the first LS dataset to be annotated with lexical complexity values using a 5-point Likert-scale and to provide gold candidate simplifications catering for all tasks within the LS pipeline. Therefore, the main contributions of this paper are:

\begin{enumerate}
    \item \textbf{MultiLS}: the first multi-purpose framework for the full training and evaluation of all LS sub-tasks (Sections \ref{related_work} to \ref{multilex_framework}).
    \item \textbf{MultiLS-PT}: the first Portuguese multi-genre dataset for LS to contain both continuous complexity values and ranked gold simplifications (Section \ref{dataset}).
    
    \item \textbf{Evaluation}: the performance of multiple state-of-the-art models for LCP, substitute generation and ranking (Sections \ref{application} to \ref{results}).

\end{enumerate}



\section{Related Work} \label{related_work}
This section establishes the need for the MultiLS framework. It introduces each task within the LS pipeline and their corresponding datasets before discussing prior end-to-end LS frameworks.

\textbf{\textit{Complexity Prediction.}} The first-step within the LS pipeline is the identification of complex words (Figure \ref{pipeline_figure}) \cite{LCPsurvey}. There are two approaches to this task. Complex Word Identification (CWI) is the first approach. CWI is a binary classification task which assigns each target word with a non-complex (0) or complex (1) label \cite{paetzold-specia:2016:SemEval1,zampieri-EtAl:2017:NLPTEA}. LCP is the second approach. It is a regression-based task (Figure \ref{pipeline_figure}). Unlike CWI, LCP is able to identify a neutral level of complexity making it the more favored approach \cite{semeval-2021}. LCP assigns a value on a continuum, including such labels as very simple (0), neutral (0.5), to very complex (1) \cite{shardlow-etal-2020-complex}. \textcolor{black}{Words that have an assigned complexity value substantially greater than 0.5 are considered to be complex words, such as the word “\textit{consultation}” within Figure \ref{pipeline_figure}.}



LCP datasets have employed the use of human annotators to assign gold complexity values. These datasets are of different languages with several being multilingual \cite{Horn2014, paetzold-specia:2016:SemEval1, yiman-EtAl:2018:BEA}. However, all of these datasets exclude Portuguese. Despite this, their annotators rate lexical complexity using a Likert-scale making them ideal for training systems to identify complex words and for substitute selection and ranking for a variety of languages. On the other hand, LCP datasets are still limited regarding their applicability. They cannot, for instance, be used to generate appropriate simplifications, referred to as candidate substitutions.






\textbf{\textit{Substitute Generation and Selection.}} Substitute generation is the second-step within the LS pipeline (Figure \ref{pipeline_figure}). Substitute generation produces candidate substitutions that are easier to understand than the original complex word while persevering its meaning \cite{north2023deep}. Substitute generation aims to produce a pre-defined number: \textit{k}, of candidate substitutions.  Substitute selection filters these generated candidates to find the  best possible simplification, commonly referred to as the top-\textit{k} candidate substitution. \textcolor{black}{For example, given the  sentence:  “\textit{Seek consultation about your diagnosis}”, and the target word: “\textit{consultation}” within Figure \ref{pipeline_figure}, substitute generation would produce \textit{k} candidate substitutions, such as “\textit{advice}”, “\textit{dialogue}”, “\textit{debate}”, and “\textit{answers}”. Substitute selection then removes those generated candidates that are  more complex, semantically dissimilar, or do not fit into the provided context resulting in the \textit{top-k} candidate substitutions: “\textit{advice}” and “\textit{answers}”.}

\textcolor{black}{Substitute generation and selection datasets provide gold candidate substitutions}. However, these datasets are independent of CWI and LCP as they do not include annotated complexity values per target word.  The ALEXSIS datasets hosted at the TSAR-2022 shared-task \cite{tsar2022} asked 25 annotators to suggest possible simplifications for English, Spanish, and Portuguese target words in context \cite{tsar2022, Ferres_Saggion2022, north2022alexsis}. The EASIER corpus \cite{Alarcn2021LexicalSS} obtained 7,892 Spanish candidate substitutions. SIMPLEX-PB 3.0 \cite{Hartmann2020} contains 3,650 Portuguese candidate substitutions and several target word features related to lexical complexity. These datasets provide data for substitute generation, selection and even substituting ranking, but do not include complexity values required for LCP.

\definecolor{Gray}{gray}{0.9}
\newcolumntype{a}{>{\columncolor{Gray}}c}
\newcolumntype{b}{>{\columncolor{Gray}}l}

\begin{table*}[!ht]
\centering

\scalebox{0.75}{\begin{tabular}{ccclcl|abab}
\multicolumn{6}{c|}{\textbf{}} &  \multicolumn{4}{a}{\textbf{MultiLS Framework (New MultiLS-PT Dataset)}} \\
\hline
\multicolumn{6}{c|}{\textbf{Original Datasets (English)}} & \textit{Step 1\textrightarrow} & \textit{Step 2\textrightarrow}& \textit{Step 3\textrightarrow}& \textit{Step 4}\\
\hline
     T. & D. & Token& Context (Sentence) & Val.& \multicolumn{1}{c|}{Substitutions} & Selection & New Context (Sentence) & New Val.& New Substitutions\\
      \hline
        \multirow{5}{*}{\rotatebox[origin=c]{90}{Task 1: LCP}}         & \multirow{5}{*}{\rotatebox[origin=c]{90}{\textbf{CompLex}}}  & colleagues & pointed out colleagues & 0.26 & \multicolumn{1}{c|}{\textbf{--}} & colegas & controlado por colegas & 0.13 & amigos (friends),.\\
        & & uncertainties & uncertainties in the & 0.37 &  \multicolumn{1}{c|}{\textbf{--}} & incertezas & influenciada por incertezas & 0.08 & dúvidas (doubts),.\\
        & & gentiles & teacher of the gentiles & 0.26 &  \multicolumn{1}{c|}{\textbf{--}} & gentios & doutor dos gentios & 0.46 & multidão (crowd),.\\
        & & prophet & raise up a prophet & 0.27 &  \multicolumn{1}{c|}{\textbf{--}} & profeta & um profeta semelhante & 0.21 & mensageiro (messenger),.\\
        & & maximum & a maximum of two & 0.14 &  \multicolumn{1}{c|}{\textbf{--}} & máximo & máximo corrigido & 0.32 & extremo (extreme),.\\

         \hline
        
         \multirow{10}{*}{\rotatebox[origin=c]{90}{Tasks 2-3: SG \& SS}} & \multirow{5}{*}{\rotatebox[origin=c]{90}{\textbf{ALEXSIS-EN}}}  & observers & the number of observers &  \multicolumn{1}{c}{\textbf{--}} & watchers, spectators,. & observadores & observadores que tiveram& 0.19 & examinadores (examiners),.\\
        & & authorities & assistance to authorities &  \multicolumn{1}{c}{\textbf{--}}  & officials, powers,. & autoridades & alegando que as autoridades & 0.23 & forças (forces),.\\
        & & condolences & sincere condolences to &  \multicolumn{1}{c}{\textbf{--}}  & sympathy, comfort,. & condolências  & suas condolências pedindos & 0.21 & compaixão (compassion),.\\
        & & regime & between Assad's regime &  \multicolumn{1}{c}{\textbf{--}}  & government, rule,. & regime & aregime do presidente & 0.11 & governo (government),.\\
        & & monitoring & it was monitoring the &  \multicolumn{1}{c}{\textbf{--}}  & watching, observing,. & monitoramento & sistema de monitoramento & 0.32 & acompanhamento,.\\

        \cline{2-10}

        & \multirow{5}{*}{\rotatebox[origin=c]{90}{\textbf{ALEXSIS+}}}  & criteria & meet the criteria &   \multicolumn{1}{c}{\textbf{--}} & requirements, standards,. & critério & critério de visão pública & 0.24 & normas (standards),.\\
        & & pledges & the agreement pledges &   \multicolumn{1}{c}{\textbf{--}} & promises, guarantees,. & promessas & faz promessas e & 0.11 & compromissos,.\\
        & & acquisition & the acquisition announced &   \multicolumn{1}{c}{\textbf{--}} & transaction, purchase,. & aquisição & local de aquisição & 0.33 & obtenção (obtaining),.\\
        & & residence & the residence next door &   \multicolumn{1}{c}{\textbf{--}} & house, apartment,. & residência & tenham residência habitual & 0.17 & casa (house),.\\
        & & inclusion & ensure the inclusion  &   \multicolumn{1}{c}{\textbf{--}} & participation, presence,. & inclusão & inclusão das opções & 0.12 & inserção (insertion),.\\

        \hline
        
        \multirow{5}{*}{\rotatebox[origin=c]{90}{Task 4: SR}} & \multirow{5}{*}{\textbf{\rotatebox[origin=c]{90}{CompLex-BC}}}  & exchange & (exchange, brains) & 1 &  \multicolumn{1}{c|}{\textbf{--}} & intercâmbio & intercâmbio efetivo das artes & 0.21 & troca (replacement),.\\
        & & sight & (sight, implants) & 0 &  \multicolumn{1}{c|}{\textbf{--}} & vista & agradável à sua vista & 0.12 & visão (view),.\\
        & & wisdom & (wisdom, women) & 1 &  \multicolumn{1}{c|}{\textbf{--}} & sabedoria & na muita sabedoria há & 0.13 & conhecimento (knowledge),.\\
        & & sword & (sword, densities) & 0 &  \multicolumn{1}{c|}{\textbf{--}} & espada &  ferimentos por espada & 0.07 & faca (knife),.\\
        & & spirit & (spirit, Mesopotamia),. & 0 &  \multicolumn{1}{c|}{\textbf{--}} & espírito & há um espírito & 0.08 & almas (souls),.\\
        
\end{tabular}}
 \caption{Illustrates the creation of MultiLS-PT. \textbf{"--"} indicates missing data in previous datasets. \textbf{T.} stands for sub-tasks within the LS pipeline that the corresponding dataset could be used for prior to MultiLS expansion. \textbf{D.} is Dataset. \textbf{Val.} represents assigned complexity value. Only a snapshot of contexts and candidate substitutions are shown.}\label{multilex_table}
\end{table*}

\textbf{\textit{Substitute Ranking.}} Substitute ranking is the final step within the LS pipeline (Figure \ref{pipeline_figure}). Substitute ranking sorts candidate substitutions from the most to least appropriate simplification. It arranges candidate substitutions based on their complexity and their semantic similarity to the target word and context \cite{north2023deep}. \textcolor{black}{The example shown in Figure \ref{pipeline_figure} ranks “\textit{answers}” as being a more appropriate simplification than “\textit{advice}” for the target word “\textit{consultation}”. This may, in part, be due to  “\textit{answers}” having a higher frequency within a reference corpus or being more frequent within a training set.  Alternatively, “\textit{answers}” may have a lower age of acquisition, higher familiarity score, or even concreteness (abstractness) rating \cite{LCPsurvey}.}


\textcolor{black}{Substitute ranking datasets allow for the comparison between candidate substitutions regarding their complexity or features correlated with complexity.} ReSyf \cite{billami-etal-2018-resyf} and HanLS \cite{HanLS}  contain 57,589 French and 534 Chinese target words respectively, ranked according to their difficulty.  ALEXSIS+ \cite{north-etal-2023-alexsis} expanded the original ALEXSIS dataset for Portuguese \cite{tsar2022} by including additional contexts and candidate substitutions annotated with cosine similarity scores. CompLex-BC \cite{north-etal-2022-evaluation} is dedicated to substitute ranking for English. It contains a subset of the CompLex dataset \cite{shardlow-etal-2020-complex} consisting of 1,940 sentences with two target words. These sentences were assigned binary comparative values of either 0 or 1; 0 indicated that candidate 1 had a greater complexity than candidate 2, and 1 denoted the opposite. A collection of other datasets contain features or readability scores useful for substitute ranking, such as SIMPLEX-PB 3.0 \cite{Hartmann2020}, the Japaneses Lexical Simplification (JLS) dataset \cite{kajiwara-yamamoto-2015-evaluation}, or the Swedish CWI dataset provided by \citet{Smolenska_2018}. However, their usefulness varies between LS sub-tasks. \textcolor{black}{Therefore, only a small amount of data currently exists that can be used for ranking Portuguese simplifications.}



\textbf{\textit{End-to-End Frameworks.}} \textcolor{black}{Previous end-to-end LS frameworks have focused on substitute generation, selection and ranking and not LCP \textcolor{black}{with no framework currently having being applied to Portuguese.} In fact, traditional notions of LS consider the identification of complex words a precursor and a separate task to LS \cite{PaetzoldSpecia2017_surveyLS}. BenchLS \cite{Paetzold2016a} provided a suitable framework for the training and evaluation of substitute generation to ranking. BenchLS \cite{Paetzold2016a} contains sentences, target words, and several candidate substitutions ranked per their simplicity, but does not supply the continuous complexity values needed for LCP. PLUMBErr \cite{Paetzold_2016_PLUMBErr}, an automatic error identification framework for LS, demonstrated its potential by assessing several LS systems that conducted complex word identification alongside all other LS sub-tasks. Nevertheless, its CWI component was trained on a dataset different from that used to evaluate its overall performance. FLELex \cite{tack-etal-2016-evaluating} does cater for LCP. It aligned two datasets of authentic and simplified texts and provided continuous complexity ratings for each target word. In this way, FLELex provides the necessary data for both LCP and all other sub-tasks of LS. However, only a proportion of their target words were labeled with a maximum of one candidate substitution per target word limiting FLELex's usefulness.}

\section{MultiLS Framework} \label{multilex_framework} %

 LS datasets often have a narrow specialization \textcolor{black}{focusing on one or two tasks.} \textcolor{black}{They only include lexical complexity values, candidate substitutions, or candidate features, restricting their use to either LCP or substitute generation, selection, or ranking} (Table \ref{multilex_table}). \textcolor{black}{Unlike previous frameworks,} the MulitLex framework supplies all the necessary data required for the training and evaluation of the entire LS pipeline, \textcolor{black}{including LCP}. \textcolor{black}{We use the MultiLS framework to  guide the creation of the first multi-purpose, and multi-genre LS dataset, named MultiLS-PT (Table \ref{multilex_table}). The MultiLS framework consists of the following summarized steps. Further details regarding MultiLS-PT are then provided in the preceding Section \ref{dataset}.}


\begin{table}[!ht]
\centering
\scalebox{0.85}{\begin{tabular}{l|l}
    \hline
\textbf{Example MTurk HIT for Annotation}  & \textbf{Difficulty} \\
      \hline
     Identify the word \textbf{\textit{\underline{authorities}}} in the sentence below:  & 1. Very Easy  \\
       &  2. Easy \\

     ``One of the greatest \textbf{\textit{\underline{authorities}}} on the subject, says    &  3. Neutral\\
       that the destruction of the biome is irreversible.'' & 4. Difficult\\
       &  5. Very Difficult\\

       \hline

\multicolumn{2}{l}{\underline{\textbf{Tasks}} } 
    \\
      
\multicolumn{2}{l}{(1). In your opinion, how difficult is the word in bold in this sentence?} \\

\multicolumn{2}{l}{\textcolor{white}{(1). }Select from 1 to 5.} \\

\multicolumn{2}{l}{(2). Write a simpler alternative to the word in bold (if any). Your suggestion} \\

\multicolumn{2}{l}{\textcolor{white}{(1). }must maintain the meaning  of the sentence above and be easier to } \\

\multicolumn{2}{l}{\textcolor{white}{(1). }understand than the word in bold. } \\
     \hline

\end{tabular}}
 \caption{An example HIT provided to the annotators. The HIT asks for both a continuous complexity rating and a suggested simplification. Each HIT was provided in Portuguese. Example has been translated for illustrative purposes.}\label{example_HIT}
\end{table}

 \begin{enumerate}
    \item \textcolor{black}{\textbf{Selection}: We identified target words from four pre-existing English datasets: CompLex \cite{shardlow-etal-2020-complex}, ALEXSIS-EN \cite{tsar2022}, ALEXSIS+ \cite{north-etal-2023-alexsis}, and CompLex-BC \cite{north-etal-2022-evaluation}. Only words with a similar use and meaning within both English and Portuguese were hand-selected to provide comparable data for future multilingual and cross-lingual experiments (Section \ref{future_work}). Selection was done by a trained linguist fluent in both languages.}
    
    \item \textcolor{black}{\textbf{Context Retrieval}: Once target words had been identified, we automatically scraped several genres (bible extracts, news articles, and biomedical papers) to obtain new and varied sentences, hereby referred to as contexts, for each target word ready for annotation.  Bible instances were \textcolor{black}{obtained} from Portuguese translations of the King James Bible. News instances were  \textcolor{black}{scraped} from the PorSimplesSent dataset \cite{leal-etal-2018-nontrivial} as well as from the CC-News (Common Crawl-News) corpus \cite{north-etal-2023-alexsis}. Biomedical instances were  \textcolor{black}{extracted} from abstracts of biomedical literature supplied by WMT-2019 \cite{bawden-etal-2019-findings}. }

    \item \textbf{New Complexities (Val.)}: We presented target words in bold within the \textcolor{black}{scraped contexts} to annotators and asked annotators to rate their perceived difficulty using a 5-point Likert-scale: very easy (1), easy (2), neutral (3), difficult (4), to very difficult (5) \cite{shardlow-etal-2020-complex, complex_2_dataset}. Each target word was annotated by 25 crowd-sourced Amazon Mechanical Turk (MTurk) workers located in Brazil. Table \ref{example_HIT} shows an example Human Intelligence Task (HIT) presented to each of the 25 annotators. \textcolor{black}{We selected a high number of annotators in order to get a representative gold complexity value for each target word by averaging the returned labels}. Annotators were paid 2 cents per annotation allowing them to surpass the minimum hourly wage in Brazil. 
    \item \textbf{New Substitutions}: Additionally, we asked annotators to suggest a valid simplification to the target word that fits within its surrounding context. Generated candidate substitutions were ranked per their suggestion frequency providing a list of gold simplifications.
\end{enumerate}



\section{MultiLS-PT Dataset}\label{dataset} %


The uniqueness of the MutliLex framework is the collection of both continuous complexity values and gold candidate substitutions. This is what gives MultiLS-PT and future datasets that follow the MultiLS framework their distinctive multi-task functionality. \textcolor{black}{The resulting MultiLS-PT dataset is unlike any other prior Portuguese dataset for LS. As referenced in Section \ref{related_work}, only two datasets exist made specifically for Portuguese LS: SIMPLEX-PB \cite{Hartmann2020}, and ALEXSIS-PT \cite{north2022alexsis}. However, these datasets only contain candidate substitutions without complexity values for target words. Moreover, both datasets are restricted to a specific genre.  MultiLS-PT, on the other hand, contains 5,165 Portuguese target words annotated with complexity values in context taken from the Bible (2,321), news articles (1,817), and biomedical texts (1,237) with each target word also having an average of two gold candidate substitutions. Table \ref{dataset_comparison} shows a direct comparison between MultiLS-PT and existing Portuguese datasets for LS.}

\begin{table}[!ht]
\centering
\scalebox{0.85}{\begin{tabular}{l|c|c|c}
    \hline
& \textbf{SIMPLEX-PB}  & \textbf{ALEXSIS-PT} & \textbf{MultiLS-PT}  \\
      \hline
     Genre & children's books & newspapers & multi-genre  \\
    \# Annotators & 5 & 25 & 25  \\
     \# Target Words & 730 & 387 & 5,165 \\
      \textbf{\# Complexity Vals.} & \textbf{-} & \textbf{-} & \textbf{5,165} \\

    \# \textbf{Substitutions} & \textbf{3,650 }& \textbf{9,605} & \textbf{9,932}  \\
     \hline
\end{tabular}}
 \caption{Comparison of Portuguese datasets for LS. MultiLS-PT is the first LS dataset to contain both gold complexity values (vals.) and candidate substitutions.}\label{dataset_comparison}
\end{table}

\section{Tasks}\label{application} 

We showcase three applications of the MultiLS-PT dataset for LS. We believed substitute selection to be conducted simultaneously during substitute generation and ranking, and therefore have only focused on LCP, substitute generation, and substitute ranking in the form of binary comparative LCP \cite{north-etal-2022-evaluation}. \textcolor{black}{Each task was defined as follows:}

 \begin{enumerate}
    \item \textcolor{black}{\textbf{Lexical Complexity Prediction (LCP)}: A regression-based task. Models were trained to automatically identify complex words by predicting their complexity value, between 0 (very easy) and 1 (very hard), of a target word in context.}

    \item \textcolor{black}{\textbf{Substitute Generation (SG)}: A text generation task. Models were set to generate top-10 (k) candidate substitutions.}
    
    \item \textcolor{black}{\textbf{Binary Comparative LCP (BC-LCP)}: A binary classification task used for substitute ranking \cite{north-etal-2022-evaluation}. Models were trained to rank candidate substitutions by assigning either 0 or 1 labels; 0 indicated that candidate 1 has a greater complexity than candidate 2 and 1 denoted the opposite.}
\end{enumerate}

\textcolor{black}{Data for each task was formatted differently for model training. Example instances with gold labels are provided below (Table \ref{example_format}). Gold labels for the three tasks were averaged complexity values, most frequently suggested simplifications, and a binary label showing which of two candidate words was more complex, respectively.}

\begin{table}[!ht]
\centering
\scalebox{0.95}{\begin{tabular}{c|l}
    \hline
     \textbf{Task} & \textbf{Example Instance with Gold Label(s)} \\ 
     \hline
\multirow{4}{*}{LCP} &  ``Procure \textbf{consulta} para diagnóstico'' <\textbackslash t> 0.73 (Gold) \\
 
&  (Translation: Seek \textbf{consultation} for diagnosis) \\
&  ``Múltiplas feridas de \textbf{espada}'' <\textbackslash t> 0.08 (Gold) \\
& (Translation: Multiple \textbf{sword} wounds)  \\
\hdashline

     \multirow{4}{*}{SG} &  ``\textbf{consulta}'' <\textbackslash t> \textbf{respostas}, \textbf{conselho}, ... (Gold)  \\

     & (Translation: \textbf{consult} <\textbackslash t> \textbf{answers}, \textbf{advice}) \\

     &``\textbf{espada}'' <\textbackslash t> \textbf{faca}, \textbf{lâmina} ... (Gold)\\

    & (Translation: \textbf{sword} <\textbackslash t> \textbf{knife}, \textbf{blade})\\

     \hdashline

    \multirow{4}{*}{BC-LCP} &  ``\textbf{respostas}'' <\textbackslash t>  ``\textbf{conselho}'' <\textbackslash t> 1 (Gold)\\

    &   (Translation: \textbf{answers} <\textbackslash t>  \textbf{advice}) \\

    &``\textbf{lâmina}'' <\textbackslash t>  ``\textbf{faca}'' <\textbackslash t> 0 (Gold)\\
    
    & (Translation: \textbf{blade} <\textbackslash t>  \textbf{knife}) \\

 \hline
\end{tabular}}
 \caption{\label{example_format} Example instances with gold labels used for training each task. Only a snapshot of gold simplifications for SG are shown. For BC-LCP, a gold label of 1 shows candidate word 1 as being less complex than candidate word 2; i.e. ``answers'' is less complex than ``advice'', whereas 0 shows the opposite.}
\end{table}

\begin{table}[!ht]
\centering
\scalebox{0.95}{\begin{tabular}{c|c|ccc|c}
    \hline
     \# & \textbf{Task} & \textbf{Train} &  \textbf{Dev} &  \textbf{Test} & \bf Total \\ 
     \hline
     1 & LCP & 3,615 & 516  & 1,034  &  5,165 \\
     2 & SG & -  & - & 462  & 462 \\  
     3 & BC-LCP & 20,113 & 2,873 & 1,029 & 24,015 \\ 
 \hline
\end{tabular}}
 \caption{\label{train_test_splits} MultiLS-PT's train, dev, and test splits per task. No training was conducted for the SG task.}
\end{table}

MultiLS-PT was divided to have a 70/10/20 corresponding train, dev, and test split for the LCP and binary comparative LCP tasks, whereas the SG task had no train, dev, and test split since it was conducted in a zero-shot setting (Table \ref{train_test_splits}). The test set of the binary comparative LCP task was also reduced by removing candidate substitution pairs that contained unrelated words and therefore were unsuitable for candidate ranking. \textcolor{black}{Each task used a different number of total instances. The LCP task leveraged all 5,165 instances. The SG and BC-LCP tasks, on the other hand, utilized smaller subsets of the MulitLex-PT dataset. The SG task used a total of 462 instances that had a minimum of 5 gold simplifications in order to conduct meaningful evaluation. The BC-LCP task used a total of 24,015 instances comparing words of similar meaning and usage per a substitute ranking scenario.}

\section{Models}\label{models} 

\begin{table*}[!ht]
\centering
\scalebox{0.85}{\begin{tabular}{ccl|p{15cm}}
\hline
   Sub-Task  & Num. & Name & Prompt \\
\hline

    \multirow{8}{*}{LCP} & 1 &  ZeroShot-5-Likert  & On a scale from 1 to 5 with 5 being the most difficult, how difficult is the "{target word}"? Answer:\\
     & 2& Context-5-Likert & On a scale from 1 to 5 with 5 being the most difficult, how difficult is the "{target word}" in the above sentence? Answer: \\
     & 3& ZeroShot-10-Likert  & On a scale from 1 to 10 with 10 being the most difficult, how difficult is the "{target word}"? Answer: \\
    & 4 & Context-10-Likert & On a scale from 1 to 10 with 10 being the most difficult, how difficult is the "{target word}" in the above sentence? Answer: \\
    & 5 & Ensemble-5-Likert & Average returned complexity from prompts 1 to 2. \\
     & 6& Ensemble-10-Likert & Average returned complexity from prompts 3 to 4.\\
     \hdashline 
     \multirow{2}{*}{SG}& 1  & ZeroShot  &  Find ten easier words in Portuguese for "{target word}". Answer: \\
     & 2& Context  &  Find ten easier words in Portuguese for "{target word}" in the above sentence. Answer:\\
     \hdashline 
     \multirow{4}{*}{BC-LCP} & 1& Difficulty  & Which word is more difficult "{target word1}" or "{target word2}"? Answer: \\
     & 2& Frequency & Which word is less common: "{target word1}" or "{target word2}"? Answer: \\
     & 3& Context & Which sentence is more difficult: (a). "{sentence1}" or (b). "{sentence2}"? Answer: \\
     & 4&  Ensemble & All of the above. \\
\hline

\end{tabular}}
 \caption{Prompts used per task.}\label{prompts}
\end{table*}



Multiple approaches using state-of-the-art models were applied to all three tasks. These approaches ranged from prompt-learning, regression, masked-language modeling (MLM) to binary classification depending on the task. Several Large Language Models (LLMs) were chosen to perform various prompt learning experiments given their high performance on a variety of NLP-related tasks. These LLMs, all of varying sizes, included GPT-3.5 (text-davinci-003) from OpenAI's API, alongside Mistral \cite{jiang2023mistral}, Llama-2 \cite{touvron2023llama}, Falcon, and MPT avialable on Hugging Face. The prompts fed into these LLMs for LCP, substitute generation, and binary comparative LCP are shown in Table \ref{prompts}. These prompts were designed to artificially replicate answers provided by human annotators by copying the instruction supplied via MTurk. 

We also experimented with several pre-trained transformers as well as a support vector machine (SVM) and a random forest (RF). Transformers and feature engineered models are currently state-of-the-art for LCP and binary comparative LCP, respectively \cite{semeval-2021, north-etal-2022-evaluation}. Transformers trained with a MLM objective were also state-of-the-art for substitute generation and selection prior to the arrival of recently proposed LLMs \cite{tsar2022, gmu-wlv-tsar-2022-shared-task}. MLM models replace the target word with a "[MASK]" special token and then attempt to provide a suitable simplification based on the masked target word and its surrounding context \cite{qiang2020BERTLS}. 


We selected several transformers pre-trained on English and/or Portuguese data. These included BERT, mBERT \cite{devlin2019bert}, RoBERTa \cite{zhuang-etal-2021-robustly}, XLM-R \cite{conneau-etal-2020-unsupervised}, BR-BERTo\footnote{\href{https://huggingface.co/rdenadai/BR_BERTo}{huggingface.co/rdenadai/BR\_BERTo}}, Albertina PT-BR\footnote{\href{https://huggingface.co/PORTULAN/albertina-ptbr}{huggingface.co/PORTULAN/albertina-ptbr}}, ALbertina PT-PT\footnote{\href{https://huggingface.co/PORTULAN/albertina-ptpt}{huggingface.co/PORTULAN/albertina-ptpt}} \cite{albertina-pt}, RoBERTa-PT-BR\footnote{\href{https://huggingface.co/josu/roberta-pt-br}{huggingface.co/josu/roberta-pt-br}}, and BERTimbau\footnote{\href{https://huggingface.co/neuralmind/bert-base-portuguese-cased}{huggingface.co/neuralmind/bert-base-portuguese-cased}} \cite{souza2020bertimbau} and were also obtained from Hugging Face. Each transformer was fine-tuned on the LCP and binary comparative LCP data supplied by MultiLS-PT as shown in Table \ref{example_format}. Fine-tuning was conducted over 5 epochs with a learning rate of 2e-5, a batch size of 8 and a max sequence length of 256 using a NVIDIA GeForce RTX 3060 GPU. No fine-tuning was conducted for substitute generation given that it is a zero-shot text generation task. Feature engineered approaches were trained on features previously shown to be indicative of lexical complexity \cite{LCP-RIT, shardlow2021predicting}. Training was conducted over 5 epochs on features ranging from word length, syllable count, frequency, prevalence, and age-of-acquisition (AoA). \textcolor{black}{Our SVM was set to have a sigmoid activation function and our RF was set to have 100 trees.} Frequencies were calculated using the Exquisite Corpus\footnote{\url{https://github.com/LuminosoInsight/exquisite-corpus}} for Portuguese. English prevalence and AoA values were taken from \citet{Brysbaertetal2019} and \citet{BrysbaertBiemiller2017}, respectively. These values were mapped to Portuguese due to the limited availability of Portuguese psycholinguistic datasets.

\textbf{\textit{Evaluation Metrics.}} Tasks were evaluated using their respective evaluation metrics found throughout LS literature \cite{vstajnerlexical}. Mean squared error (MSE), Pearson Correlation (R) and Spearman Correlation ($\rho$) were used to evaluate LCP, with lower MSE values correlated with greater performance \cite{semeval-2021}. Weighted average recall, precision, and F1-score were used to assess binary comparative LCP \cite{north-etal-2022-evaluation}. However, substitute generation was evaluated using a alternative set of evaluation metrics introduced in the TSAR-2022 shared-task \cite{vstajnerlexical, tsar2022}, including potential and accuracy at top-\textit{k} = 1.  Potential is the ratio of the predicted candidate substitutions that match the most frequently suggested gold label. Accuracy at top-\textit{k} = 1 (A@1@Top1) is the ratio of best predicted candidate substitutions at rank \#1 that are equal to the most appropriate gold simplification also at rank \#1. It is important to note, A@1@Top1 is different from ACC@1 that is reported alongside A@1@Top1 at TSAR-2022 \cite{tsar2022}. ACC@1 takes into consideration multiple generated candidates, whereas A@1@Top1 only considers the top-\textit{k} = 1 candidate generated. We decided to use A@1@Top1 as it is a more competitive evaluation metric.



\section{Results}\label{results} 

\begin{table*}[!ht]
\centering
\scalebox{0.85}{\begin{tabular}{ccl|ccc|ccc|ccc|ccc}
\hline
   & & & \multicolumn{3}{c|}{\textbf{All}} & \multicolumn{3}{c|}{\textbf{Bible}} & \multicolumn{3}{c|}{\textbf{News}} & \multicolumn{3}{c}{\textbf{Biomed}} \\
\hline
       Approach &  \# & Model& MSE &  R & $\rho$ & MSE &  R & $\rho$ & MSE &  R & $\rho$ & MSE & R & $\rho$\\
       \hline
   \multirow{7}{*}{Transformers}   & 1  &  BERTimbau  &  0.0664  &  0.8423  &  0.8081  &  0.0726  &  0.8260  &  0.8275  &  0.0558  &  0.7244  &  0.7047  &  0.0677  &  0.8959  &  0.8740  \\
      & 2  &  BERTimbau-L &  0.0681  &  0.8324  &  0.8086  &  0.0746  &  0.8144  &  0.8227  &  0.0533  &  0.7450  &  0.7308  &  0.0746  &  0.8720  &  0.8573 \\
      & 3  &  XLM-R-L &  0.0698  &  0.8295  &  0.8054  &  0.0777  &  0.8055  &  0.8224  &  0.0550  &  0.7212  &  0.7214  &  0.0724  &  0.8907  &  0.8586  \\
      & 4  &  XLM-R  &  0.0706  &  0.8187  &  0.7974  &  0.0773  &  0.8012  &  0.8155  &  0.0595  &  0.6774  &  0.6995  &  0.0716  &  0.8824  &  0.8612 \\
      & 5  &  mBERT  &  0.0743  &  0.7968  &  0.7724  &  0.0815  &  0.7746  &  0.7808  &  0.0585  &  0.6801  &  0.6941  &  0.0804  &  0.8502  &  0.8332 \\
      & 6  &  RoBERTa-PT-BR  &  0.1469  &  0.7968  &  0.7539  &  0.1506  &  0.7440  &  0.7395  &  0.1169  &  0.7214  &  0.6834  &  0.1811  &  0.8768  &  0.8430 \\
      & 7  &  BR-BERTo  &  0.1844  &  0.7522  &  0.6791  &  0.1842  &  0.6865  &  0.6500  &  0.1488  &  0.6518  &  0.5906  &  0.2340  &  0.8569  &  0.8111 \\
      \hdashline
     \multirow{4}{*}{LLMs} & 8 & \textcolor{black}{Mistral-8X7B} & 0.1810 & 0.4603 & 0.4816 & 0.1576 & 0.5608 & 0.5480 & 0.1953 & 0.4663 & 0.4566 & 0.2063 & 0.3762 & 0.3877 \\

     & 9 & \textcolor{black}{Llama-2-13B} & 0.2249 & 0.2737 & 0.3441 & 0.2089 & 0.2226 & 0.3330 & 0.2289 & 0.2569 & 0.3233 & 0.2535 & 0.2687 & 0.2903 \\

     & 10 &Mistral-7B & 0.4156 & 0.2758 & 0.3349 & 0.4117 & 0.3762 & 0.3880 & 0.4428 & 0.3379 & 0.3327 & 0.3739 & 0.1148 & 0.2261 \\
     & 11  &  GPT 3.5 &  0.5050  &  0.0520  &  0.0895  &  0.5019  &  0.0134  &  0.0481  &  0.5286  &  0.0624  &  0.1197  &  0.4692  &  0.1411  &  0.1504 \\
     &  12  & Llama-2-7B  &  0.4031  &  0.0392  &  0.1535  &  0.4064  &  0.0394  &  0.1343  &  0.4199  &  0.1951  &  0.2084  &  0.3631  &  -0.0121  &  0.1287 \\
     &  13  & Falcon-7B  &  0.4273  &  0.0008  &  0.0353  &  0.4150  &  -0.019  &  -0.0132  &  0.4722  &  0.0718  &  0.0993  &  0.3703  &  0.0285  &  0.0613 \\


     
    \hline
       \multicolumn{15}{c}{\textbf{LCP-2021 Benchmark (English)}} \\
    \hline

    Transformers & 1 & BERT-Ensemble &  0.0609 & 0.7886 & 0.7369 & - & - & - & - & - & - & - & - & -  \\

\hline
\end{tabular}}
 \caption{LCP performances on instances separated by genre. Models are ranked (\#) from best to worst Pearson Correlation (R) for all instances. Results produced by LLMs were from our highest performing prompt 1. ZeroShot-5-Likert (Table \ref{prompts}). \textcolor{black}{The winning system from LCP-2021 \cite{semeval-2021} provided as a benchmark}.}\label{results_lcp}
\end{table*}

\textcolor{black}{The following section details our results for each task using the MultiLS-PT dataset. We report model performances on LCP (Table \ref{results_lcp}) before moving to substitution generation (Table \ref{results_SG}), and finally substitute ranking via binary comparative LCP (Table \ref{resultsSG}). For each task, we look into LLM versus transformer performance, impact of genre and context, and compare model performances on MultiLS-PT to prior datasets. Section \ref{discussion} discusses these areas in more detail in relation to the entire LS pipeline and provides several conclusions.}

\textcolor{black}{\textbf{\textit{Lexical Complexity Prediction (LCP).}} Pre-trained transformers outperformed our LLMs for LCP, regardless of genre or prompt (Table \ref{results_lcp}). Transformers fine-tuned on all of the instances from MultiLS-PT, depicted lower MSE values alongside higher R and $\rho$ values compared with our prompt learning approaches. The highest performing models were BERTimbau (\#1) and XLM-R-L (\#3) having achieved R values of 0.8423 and 0.8295,  $\rho$ values of 0.8081 and 0.8054, and MSE values of 0.0664 and 0.0698, respectively. In comparison, our best performing LLMs achieved noticeably worst performances when asked to rate the complexity of the target word in a zero-shot setting (ZeroShot-5-Likert, Table \ref{prompts}). Mistral-8X7B (\#8) achieved a R value of 0.1810, a $\rho$ value o 0.4816, and a MSE value of 0.1810. Llama-2-13B (\#13) produced a R value of 0.2249, a $\rho$ value o 0.3441, and a MSE value of 0.2249. All other prompts that took into consideration context or had their answers averaged within an ensemble resulted in worst performances. This was due to the nature of the task. Without prior exposure to gold complexity ratings, our prompts were ineffective at modeling the complexity assignments of Portuguese speakers.} 

\textcolor{black}{Differences in LCP performance per genre were observed by both transformers and LLMs. Transformers fine-tuned and evaluated on biomed instances returned the best results followed by Bible and news extracts. BERTimbau (\#1) produced R values of 0.8959, 0.8260, and 0.7244 on biomed, Bible, and news instances, respectively. Likewise, XLM-R-L (\#3) achieved R values of 0.8907, 0.8055, and 0.0.7212 on biomed, Bible, and news instances, respectively. Interestingly, Mistral-8x7B performed best on Bible instances having achieved a R value of 0.5608, followed by news instances attaining a R value of 0.4663, and lastly biomedical instances scoring a R value of 0.3762. Varying performances between genre can be seen throughout the remaining tasks.}

\begin{table*}[!ht]
\centering
\scalebox{0.85}{\begin{tabular}{cp{3.5cm}|cc|cc|cc|cc}
\hline
  & &  \multicolumn{2}{c|}{\textbf{All}} & \multicolumn{2}{c|}{\textbf{Bible}} & \multicolumn{2}{c|}{\textbf{News}} & \multicolumn{2}{c}{\textbf{Biomed}} \\
\hline
    \# & Model-Prompt & A@1@Top1  & Potential & A@1@Top1 & Potential & A@1@Top1 & Potential & A@1@Top1  & Potential\\
      \hline

       1 & \textcolor{black}{Falcon-40B-Context}  & 0.1708 & 0.5291 & 0.2086 & 0.5043 & 0.1329 & 0.5606 & 0.1428 & 0.5324 \\
               
       2 & \textcolor{black}{Falcon-40B-ZeroShot} & 0.1375 & 0.4333 & 0.1826 & 0.4521 & 0.0867 & 0.4219 & 0.1168 & 0.4025 \\

       3 & \textcolor{black}{Mistral-8x7B-Context} & 0.1375 & 0.4083 & 0.1695 & 0.4173  & 0.1040 & 0.3988  & 0.1168 & 0.4025   \\

       4 & \textcolor{black}{Mistral-8x7B-ZeroShot} & 0.1125 & 0.3187  & 0.1260 & 0.2739 & 0.0693 & 0.3294 & 0.1688 & 0.4285  \\  

      5 & \textcolor{black}{Llama-2-13B-Context} & 0.1104 & 0.3208 & 0.1260 & 0.2869 & 0.0867 & 0.3526 & 0.1168 & 0.3506 \\

        6   &   GPT 3.5-ZeroShot  &   0.1083  &  0.3479  &  0.1217  &  0.3043  &  0.0867  &  0.3930  &  0.1168  &  0.3766 \\
        
        7   &  GPT 3.5-Context  &  0.1062  &  0.4250  &  0.1304  &  0.3956  &  0.0693  &  0.4797  &  0.1168  &  0.3896 \\

        8  &  Mistral-7B-Context  &   0.0937  &  0.2750  &  0.1086  &  0.2652  &  0.0693  &  0.2716   &  0.1038  &  0.3116 \\

        9   &   BERTimbau  &   0.0916  &  0.2645  &  0.1086  &  0.2434  &  0.0751  &  0.2890  &  0.0779  &  0.2727 \\

        10  &  Mistral-7B-ZeroShot  &  0.0729  &  0.2062  &  0.0826  &  0.1913  &  0.0578  &  0.2080  &  0.0779  &  0.2467 \\

       11 & \textcolor{black}{Llama-2-13B-ZeroShot} & 0.0520 & 0.1187 & 0.0739 & 0.1565 & 0.0231 & 0.0809 & 0.0519 & 0.0909   \\

        12   &  XLM-R  &   0.0458  &  0.1250  &  0.0478  &  0.0913  &  0.0462  &  0.1618  &  0.0389  &  0.1428 \\

        13   &   RoBERTa-PT-BR  &   0.0333  &  0.1229  &  0.0434  &  0.1000  &  0.0173  &  0.1329  &  0.0389  &  0.1688 \\

        14   &   mBERT   &   0.0229  &  0.1145  &  0.0217  &  0.0826  &  0.0231  &  0.1445  &  0.0259  &  0.1428 \\

        15   &   Llama-2-7B-ZeroShot  &   0.0229  &  0.1000  &  0.026  &  0.0782  &  0.0115  &  0.1213  &  0.0389  &  0.1168 \\

        16   &   MPT-7B-Context  &  0.0229  &  0.0958  &  0.0260  &  0.0826  &  0.0115  &  0.1040  &  0.0389  &  0.1168 \\

        17   &   BR-BERTo  &   0.0250  &  0.0770  &  0.0304  &  0.0391  &  0.0173  &  0.1098  &  0.0259  &  0.1168 \\

        18   &   MPT-7B-ZeroShot  &  0.0208  &  0.0750  &  0.0260  &  0.0652  &  0.0057  &  0.0867  &  0.0389  &  0.0779 \\

        19   &   Llama-2-7B-Context  &   0.0208  &  0.0541  &  0.0260  &  0.0391  &  0.0057  &  0.0635  &  0.0389  &  0.0779 \\

        20   &   Falcon-7B-Context  &  0.0166  &  0.0416  &  0.0173  &  0.0478  &  0.0057  &  0.0346  &  0.0389  &  0.0389 \\

        21  &   Albertina PT-BR  &   0.0145  &  0.0541  &  0.0173  &  0.0478  &  0.0115  &  0.0635  &  0.0129  &  0.0519 \\

        22   &   Albertina PT-PT &   0.0145  &  0.0520  &  0.0173  &  0.0478  &  0.0115  &  0.0578  &  0.0129  &  0.0519 \\

      \hline
       \multicolumn{10}{c}{\textbf{TSAR-2022 Benchmark (PT-BR)}} \\
      \hline
         1  &  BERTimbau  & - & - & - & - & 0.2540 & 0.4812 &  - & - \\
        
\hline
\end{tabular}}
 \caption{Shows substitute generation performances on instances separated by genre with at least five gold candidate substitutions in MultiLS-PT. Models are ranked (\#) from best to worst A@1@Top1. LLMs are separated by inputted prompt. \textcolor{black}{The winning system from TSAR-2022 \cite{tsar2022} provided as a benchmark.}}\label{results_SG}
\end{table*}

\textbf{\textit{Substitute Generation.}} \textcolor{black}{Simplifications generated by  LLMs were of a greater quality compared to those generated by the majority of MLM approaches for all instances (Table \ref{results_SG}). The best LLM, being Falcon-40B (\#1), achieved an A@1@Top1 of 0.01708 and a potential of 0.5291, closely followed by Mistral-8x7B (\#3) having obtained an A@1@Top1 of 0.1375, and a potential of 0.4083. (Table \ref{results_SG}). The majority of MLM approaches, including transformers such as XLM-R (\#12), RoBERTa-PT-BR (\#13), mBERT (\#14), and so on, produced less suitable candidate substitutions with A@1@Top1 scores of 0.0458, 0.0333, 0.0229, respectively. However, the best performing MLM model, being BERTimbau (\#9), achieved an A@1@Top1 of 0.0916, that surpassed the performance of smaller LLMs, such as Mistral-7B, and Llama-2-7B. A direct correlation was therefore observed between LLM size and overall performance.} 

\textcolor{black}{Context influenced prompt performance. The three best performing LLMs, Falcon-40B (\#1), Mistral-8x7B (\#3), and Llama-2-13B (\#5), produced their best simplifications across all genres when fed prompts referring to the target word's context. Their Zero-shot counterparts, on the other hand, performed noticeably worst. Falcon-40B scored a A@1@Top1 of 0.1708 with context dropping to 0.1375 without context. Mistral-8x7B achieved a A@1@Top1 of 0.1375 with context falling to 0.1125 without context.  Llama-2-13B showed the greatest decrease in performance having fell from a A@1@Top1 of 0.1104 with context to a much lower A@1@Top1 of 0.0520 without context. This signifies the vital role context plays in substitute generation.} 

\textcolor{black}{Substitute generation performance also varied between genres. Falcon-40B (\#1), Mistral-8x7B (\#3), and Llama-2-13B (\#5) achieved greater A@1@Top1 and potential scores for Bible instances when compared to biomed and news instances. Falcon-40B  Mistral-8x7B, and Llama-2-13B  produced candidate substitutions with A@1@Top1 scores of 0.2086, 0.1695, and 0.1260 for Bible instances, respectively. However, the same LLMs produced inferior candidate substitutions for news extracts with A@1@Top1 scores of 0.1329 by Falcon-40B, 0.1040 by Mistral-8x7B, and 0.0867 by Llama-2-13B. Little variation, however, was seen between these LLMs performance on the biomedical extracts with Mistral-8x7B and Llama-2-13B achieving the same A@1@Top1 of 0.1168.}

\textcolor{black}{Performances on the news genre were lower than those achieved at the TSAR-2022 shared-task \cite{tsar2022}. The wining system of TSAR-2022's Portuguese track was an BERTimbau-based system that achieved an A@1@Top1 of 0.2540 on the shared-task's news extracts \cite{gmu-wlv-tsar-2022-shared-task}. Our best performing model, being Falcon-40B (\#1), achieved an A@1@Top1 of 0.1329 for news instances. We attribute this performance to how MultiLS-PT's news instances were collected. Target words within MultiLS-PT's news genre were taken from CompLex's European Parliamentary proceedings (Parl) genre \cite{shardlow-etal-2020-complex}. This was done to maintain a level of similarity between the  two datasets as described in Section \ref{dataset}. However, as a consequence, this resulted in more nuanced and complex sentences being present among MultiLS-PT's news instances in comparison to TSAR-2022's news extracts making substitute generation a more challenging task.}






\begin{table}[!ht]
\centering
\scalebox{0.85}{\begin{tabular}{cl|cccc}
\hline
   & &  \multicolumn{4}{c}{\textbf{F1-Score}}\\
\hline
        \# & Model-Prompt/Features&   \multicolumn{1}{c|}{\textbf{All}} & \multicolumn{1}{c|}{\textbf{Bible}} & \multicolumn{1}{c|}{\textbf{News}} & \multicolumn{1}{c}{\textbf{Biomed}} \\
      \hline
       1  &  GPT 3.5-Frequency &  0.7064 & 0.6555 & 0.7474 & 0.6063 \\
       
        2 & \textcolor{black}{Mistral-8x7B-Frequency} & 0.6992 & 0.5907 & 0.6986 & 0.6087\\
       
       3  &  Mistral-7B-Difficulty &  0.6276 & 0.6556 & 0.5989 & 0.6516 \\



       4  &  Llama-2-7B-Ensemble &  0.6015 & 0.6168 & 0.5826 & 0.5984 \\
       5  &  mBERT &  0.5223 & 1.0000 & 0.5932 & 0.3213 \\
       6  &  Falcon-7B-Frequency &  0.5097 & 0.4771 & 0.2536 & 0.4781 \\
       7  &  RF-all &  0.5044 & 0.5472 & 0.4938 & 0.3999 \\

       8  & \textcolor{black}{Llama-2-13B-Difficulty} & 0.5043 & 0.5225 & 0.6493 & 0.5986\\

       9  &  SVM-all &  0.4995 & 0.5030 & 0.4721 & 0.4875 \\
       10  &  MPT-7B-Difficulty &  0.4789 & 0.5212 & 0.5633 & 0.5085 \\

       11 & \textcolor{black}{Flacon-40B-Sentence} & 0.4737 & 0.4692 & 0.4684 & \textcolor{black}{0.6355}\\
       
       13  &  XLM-R &  0.4434 & 0.3995 & 0.4111 & 0.4579 \\

  \hline
  
\end{tabular}}
 \caption{Shows weighted average binary comparative LCP F1-scores on instances separated by genre and language. Performances are shown as weighted averages. Models are ranked (\#) from best to worst F1-Score for all instances. LLMs are separated by inputted prompt.}\label{resultsSG}
\end{table}

\textbf{\textit{Binary Comparative LCP.}} \textcolor{black}{GPT 3.5 achieved the best performance for binary comparative LCP. For the majority of instances, GPT 3.5 (\#1) and Mistral-8x7B (\#2) were able to predict which of two target words were more or less complex having achieved F1-scores of 0.7064 and 0.6992 for all instances, respectively. Unlike for LCP, no clear distinction was observed between the performances of several LLMs and transformers. For example, mBERT achieved an F1-score of 0.5223, whereas other larger LLMs, such as Llama-2-13B (\#8), Falcon-7B (\#6) and Falcon-40B (\#11) attained F1-scores of 0.5043, 0.5097, 0.4737, respectively. This was likely due to the difficult nature of the task as discussed within Section \ref{discussion}.}

\textcolor{black}{Features known to correlate with complexity were embedded within several prompts to better understand the thought process of our LLMs (Table \ref{prompts}). It was discovered that LLMs performed differently when taking into consideration different prompts. GPT 3.5 (\#1) and Mistral-8x7B (\#2) achieved their greatest F1-scores of 0.7064 and 0.6992 respectively when being asked to determine which target word was more or less complex based on its frequency. In contrast, these same models achieved noticeably worst F1-scores when being fed prompts that explicitly referred to word difficulty (Table \ref{prompts}). When inputted difficulty-based prompts, GPT 3.5 produced a F1-score of 0.6273 and Mistral-8x7B achieved a F1-score of 0.6154 amounting to a -0.0791 and -0.0838 decrease in performance respectively. Therefore, it would appear that our best performing LLMs considered frequency as being a highly influential factor in determining a word's overall complexity.}



\textcolor{black}{On several occasions, prompt performance varied between genres for binary comparative LCP.  GPT 3.5 (\#1) and Mistral-8x7B (\#2) were able to use frequency-based prompts to differentiate the complexities of words taken from the news genre more easily than they were for words taken from the Bible or biomed genres. For the news genre, GPT 3.5 attained a F1-score of 0.7474 and Mistral-8x7B produced a F1-score of 0.6986. However, for the Bible and biomed genre, GPT 3.5 produced F1-scores of 0.6555 and 0.6063 respectively, whereas as Mistral-8x7B achieved F1-scores of 0.5907 and 0.6087 respectively. Interestingly, Falcon-40B (\#12) produced it's highest F1-score when using sentence-based prompts (Table \ref{prompts}) for ranking words from the biomed genre. A probable explanation likely stems from the varying lexical diversity of each genre. The news genre was found to contain a greater combination of everyday and jargon-specific vocabulary making its complex and non-complex words easier to differentiate. The vocabulary of the Bible and biomed genres, on the other hand, were more jargon-specific making binary comparative LCP a harder task when considering word frequency, yet an easier task when comparing two target sentences since surrounding words are also taken into consideration. }





\section{Discussion} \label{discussion}

\textcolor{black}{This section discusses task, genre, and context performance in relation to the entire LS pipeline. We begin by giving a recommended pipeline per our results before discussing discrepancies between genre performances and the impact context plays in LS. }

\textbf{\textit{Recommended Pipeline.}}\label{pipeline_discussion} The MultiLS framework creates novel LS datasets that are capable of developing an entire end-to-end LS pipeline, \textcolor{black}{including LCP}. From the experiments carried out on MultiLS-PT (Section \ref{results}), we were able to determine the best approaches for LCP, substitute generation, and binary comparative LCP. We can, therefore, theorize \textcolor{black}{the current optimum LS pipeline for Portuguese}.

Transformers trained for regression maintained their status as state-of-the-art for LCP, being the first task within the LS pipeline \cite{semeval-2021}. This is likely due to our transformers being fine-tuned on the MultiLS-PT dataset. Lexical complexity is a highly subjective and intrinsic task. Assigned complexity values have been shown to vary from annotator-to-annotator \cite{maddela2018word, North_frontiers}. The poor performance of LLMs at LCP is therefore unsurprising. A deeper understanding of how a specific target demographic rates lexical complexity is required. In addition, the use of a single LLM to replicate complexity assignment may be considered unfair, since multiple annotators were tasked with assigning gold labels. With this in mind, we averaged the complexity assignments of several LLMs (Table \ref{comparison_LCP}). A slight increase in performance was observed. However, performances were still inferior in comparison to our fine-tuned transformers.

\begin{table}[!ht]
\centering
\scalebox{0.85}{\begin{tabular}{ccl|ccc}
\hline
        Lang.  & Model& MSE &  R & P \\
      \hline
     \multirow{2}{*}{EN}   &Mist-Llama-7B & 0.2728 & \textbf{0.2120} & 0.2853 \\
      &Mist-Llama-GPT-7B & 0.3051 & 0.1644 & 0.2210  \\
    \hdashline
     Baseline  &Mistral-7B & 0.2490 & 0.1933 & 0.2327 \\ 
    \hline
         \multirow{2}{*}{PT}  &Mist-Llama-7B & 0.4010 & 0.2116 & 0.3242  \\
    &Mist-Llama-GPT-7B & 0.4088 & 0.1406 & 0.2190  \\
     \hdashline
    Baseline & Mistral-7B & 0.4156 & \textbf{0.2758} & 0.3349 \\
 \hline
\end{tabular}}
 \caption{LCP performances on \textbf{all} instances for LLM ensembles. Models are ranked (\#) from best to worst Pearson Correlation (R).}\label{comparison_LCP}
\end{table}

For the second task of substitute generation, LLMs that were fed context-orientated prompts produced the best simplifications compared to transformers with a MLM objective. Consistent with the findings of TSAR-2022 \cite{tsar2022}, the candidate substitutions produced by a LLM had a higher likelihood of matching the gold simplifications provided by the annotators of MultiLS-PT. This was especially true if the context of the original target word was included within the LLM's prompt. 

The final task of the LS pipeline, being substitute ranking, saw GPT 3.5 and Mistral-8x7B correctly sort the majority of instances when making decisions based on word frequency. Ranking via binary comparative LCP is a challenging task. Words that at first glance appear to be complex may in fact be non-complex and vice-versa. This makes ranking candidate substitutions notoriously difficult \cite{north-etal-2022-evaluation}. As such, larger models, such as GPT 3.5 (text-davinci-003) and Mistral-8X7B, were found to perform exceptionally well at binary comparative LCP when being fed frequency-based prompts. However, this did not apply to all 7B plus parameter models as Falcon-40B and Llama-2-13B were found to perform poorly. This was likely a results of the data used to train each individual LLM with some having more exposure to Portuguese than others.

Our experiments on the MultiLS dataset have allowed us to theorize the \textcolor{black}{optimum} LS pipeline for Portuguese given current state-of-the-art models: 

\begin{enumerate}
    \item \textbf{Complexity Prediction}: A pre-trained transformer fine-tuned for LCP regression.
    \item \textbf{Substitute Generation}: A LLM provided with a context-orientated prompt.
    \item \textbf{Substitute Ranking}: GPT 3.5 or Mistral over 7B parameters fed with frequency-based prompts.
\end{enumerate}

\textbf{\textit{Genre and LS.}}\label{genre_discussion} The MultiLS framework is the first of its kind to allow for the exploration of the effects of genre on all sub-tasks within the LS pipeline. Discrepancies were found between the performance of each task and genre. Performances were greater for LCP and substitute generation on jargon-rich genres, such as biomedical articles or European Parliamentary proceedings, or that contained more archaic vocabulary, such as the Bible. For example, target words, such as "\textit{correlação}" (correlation) or "\textit{hipertensão}" (hypertension) from biomedical articles, or "\textit{profecia}" (prophecy) or "\textit{myrrh}" from the Bible were found to be easier to identify as complex and had fewer suitable simplifications, increasing the likelihood of candidate substitutions matching the gold simplification at \textit{k}=1. Binary comparative LCP, on the other hand, benefited from the vernacular of newspaper articles. Ranking candidate substitutions such as  "\textit{ruas}" (streets) or "\textit{vias}" (routes) to replace \textit{estradas} (roads), or "\textit{problema}" (problem) or "\textit{conflito}" (conflict) to replace \textit{crise} (crisis) were found to be  easier for GPT 3.5 and Mistral-8X7B than compared to ranking potential simplifications for the aforementioned medical or biblical jargon. With this in mind, an LS pipeline would benefit from genre fine-tuning rather than relying on models pre-trained on multiple genres.



\textbf{\textit{Context and LS.}}\label{context_discussion} The MultiLS framework provides annotators with target words in context. Experimentation on a MultiLS dataset has proven that context is a deciding factor for LCP and substitute generation, but not necessarily for binary comparative LCP. The complexity of surrounding words influences the perceived lexical complexity of a target word and therefore is vital for complexity prediction \cite{LCPsurvey}. Context must also be known in order to provide a valid simplification making it highly important for substitute generation \cite{gmu-wlv-tsar-2022-shared-task}. However, the comparison of candidate substitutions embedded within related sentences is an ineffective means of substitute ranking. In most cases, such contextual comparisons were found to confuse our binary comparative LCP models and was only effective when employed be an exponentially large LLM (Falcon-40B) for a specific genre: biomed. This is because a target word's context is not always representative of a target word's true complexity.




\section{Conclusion}

The MultiLS framework provides a guide for the creation of a multi-purpose and multi-genre LS dataset. The MultiLS framework is unique in that it provides gold continuous complexity values and gold candidate substitutions, a feat not achieved by previous LS datasets (Sections \ref{related_work} and \ref{dataset}). The resulting dataset can be used to train and evaluate all LS sub-task, \textcolor{black}{including LCP}. 


We introduce MultiLS-PT, the first Portuguese LS dataset to be created using the MultiLS framework. By experimenting on MultiLS-PT, we were able to theorize the \textcolor{black}{optimum LS pipeline for Portuguese given current state-of-the-art models} and make several observations regarding the impact of genre and context on LS. Performances indicate that LLMs are incapable of rating lexical complexity for a specific target demographic, but are able to generate and rank possible simplifications. This provides insight into the role LLMs will have in future LS systems.


\section{Future Work} \label{future_work}

In this paper, we provided empirical evidence of the MultiLS framework's potential to be used as an all-in-one simplification framework. We have trained models and conducted several experiments using MultiLS-PT. However, there are multiple research questions left outstanding that the MultiLS framework and MultiLS-PT can be used to answer. Future work will utilize the MultiLS framework to explore three open research areas within the field of LS: (1). full pipeline evaluation, (2). multilingual LS and cross-lingual transfer, and (3). domain generalization \cite{north2023deep}. 
 
\textbf{\textit{Full Pipeline Evaluation.}} LLMs are able to simplify an entire text as a response to a single prompt and are even state-of-the-art for substitute generation (Section \ref{results}). This questions the need for models trained on individual sub-tasks of the LS pipeline. Comparisons need to be made between the readability and accessibility of texts simplified by a general LLM compared to texts simplified by end-to-end LS systems. To make this possible, we aim to compare the performance of an LS pipeline trained on a MultiLS dataset to a generalized LLM for text simplification.

\textbf{\textit{Multilingual LS and Cross-lingual Transfer.}} The number of LS datasets varies between languages.  Languages, such as English, Spanish, French, Japanese, and Chinese, are relatively well-resourced for LS \cite{merejildo2021, billami-etal-2018-resyf, kajiwara-yamamoto-2015-evaluation, Lee_Yeung2018}. However, other languages, including Bengali, Russian, Swedish, and others, are under-resourced \cite{Abramov_2022, Smolenska_2018}.  Training a LS system on a well-resourced language and then applying the same LS system to a low-resource language may be a successful means of improving LS performance without the need of collecting new LS datasets. However, conflicting evidence exists regarding the performance of cross-lingual models for LS \cite{gmu-wlv-tsar-2022-shared-task, vstajnerlexical,  North_frontiers}. Further research is needed to establish whether cross-lingual transfer is viable for LS, especially for which LS sub-tasks. In this endeavour, we plan to apply the MultiLS framework to other languages whereby the complexities of shared and hand-selected words can be used to research the effects of multilingual LS and cross-lingual transfer on LS performance. 

\textbf{\textit{Domain Generalization.}} LS systems are commonly trained on a single dataset containing either a specific genre, including newspaper extracts \cite{Aluisio2010, paetzold-specia:2016:SemEval1, leal-etal-2018-nontrivial, north2022alexsis} or educational materials \cite{Hartmann2020, Zambrano2020OverviewOA, merejildo2021}, or for an undefined mix of genres, such as Wikipedia extracts on a range of topics \cite{CWcorpus, Horn2014}. The lack of datasets containing multiple types of text separated by genre limits the development of LS systems capable of domain generalization. The MultiLS framework controls genre. As such, researchers can see what does and does not work well for specific genres and use this information to develop their LS systems accordingly.  Using the findings of this paper, we aim to continue to experiment with MultiLS-PT developing a fully generalizable LS system for Portuguese.



\bibliographystyle{ACM-Reference-Format}
\bibliography{CWI}

\appendix

\end{document}